# A computational model of early language acquisition from audiovisual experiences of young infants

*Okko Räsänen*[1,2], *Khazar Khorrami*[1]

[1]Faculty of Information Technology and Communication Sciences, Tampere University, Finland
[2]Department of Signal Processing and Acoustics, Aalto University, Finland
`okko.rasanen@tuni.fi, khazar.khorrami@tuni.fi`

## Abstract

Earlier research has suggested that human infants might use statistical dependencies between speech and non-linguistic multimodal input to bootstrap their language learning before they know how to segment words from running speech. However, feasibility of this hypothesis in terms of real-world infant experiences has remained unclear. This paper presents a step towards a more realistic test of the multimodal bootstrapping hypothesis by describing a neural network model that can learn word segments and their meanings from referentially ambiguous acoustic input. The model is tested on recordings of real infant-caregiver interactions using utterance-level labels for concrete visual objects that were attended by the infant when caregiver spoke an utterance containing the name of the object, and using random visual labels for utterances during absence of attention. The results show that beginnings of lexical knowledge may indeed emerge from individually ambiguous learning scenarios. In addition, the hidden layers of the network show gradually increasing selectivity to phonetic categories as a function of layer depth, resembling models trained for phone recognition in a supervised manner.

**Index Terms**: language acquisition, computational modeling, L1 acquisition, lexical learning, phonetic learning

## 1. Introduction

Despite decades of research, it is still unclear how infants learn to understand speech through their interactions with the surrounding world and linguistic community. Key questions include 1) how infants succeed in initial segmentation of words from running speech, 2) how they learn to map the words to their referential meanings (cf. [1]), and 3) whether the emerging phonemic knowledge of the language precedes, parallels, or follows from early lexical learning. The existing research has come up with several suggestions of how infants may solve these different tasks, such as using statistical regularities (e.g., [2]), prosody [3], or other properties infant-directed speech [4]. However, it is unclear what type of mechanisms are actually feasible with real-world sensory experiences of language learning infants, i.e., with input that often differs radically from laboratory experiments. In addition, it is unclear how the different subtasks in learning interact and depend on each other.

Computational modeling, on the other hand, enables evaluation of theories and models while considering multiple aspects of the learning process simultaneously (see [5] for a recent overview). In this respect, previous models have shown that fully unsupervised acoustic word discovery is challenging due to the variable nature of acoustic speech, even though some recurrent patterns are still learnable (e.g., [6–10]). From machine learning point of view, the main issue is that fully unsupervised learning of linguistic patterns is an ill-posed problem, necessarily requiring several built-in assumptions on the type of patterns that are being searched for. In contrast, if speech heard by the learner is paired with visual stimuli that correlate with nouns and verbs in the utterances, learning of the object and action words becomes a weakly-supervised learning problem with a well-defined learning criterion (Section 2). By treating concurrent visual input as noisy and poorly aligned labeling for speech, statistical learning across multiple individually ambiguous interaction scenarios can help to resolve how acoustic patterns relate to the external world.

To follow such an idea, several computational models of LA have utilized the idea of concurrent audiovisual learning from acoustic speech input. Pioneering models such as CELL by Roy and Pentland [12] and that of Yu and Ballard [13] were followed by others, such as visually-conditioned higher-order Markov chains [11, 14, 15], audiovisual non-negative matrix factorization [16, 17] (see also [18, 19] for related work) and DP-n-grams [20]. These models have demonstrated successful word learning from acoustic speech when the speech comes with related visual information, provided as an unaligned bag of categorical labels for each utterance. Similar ideas have been recently re-emerged in the area of low-resource speech retrieval applications where models are trained with images and their read-aloud text captions (e.g., [21, 22]). Another branch of studies has focused on audiovisual learning of word meanings from referentially ambiguous learning scenarios by using pre-segmented and categorical word representations (e.g., [23–26]).

However, according to our knowledge, none of the existing models for joint segmentation, meaning acquisition, and sub-word unit learning have been tested with real naturalistic input available to language learning infants (see also [5]). Instead, the models (or robots) have used supervised phone recognizer front-ends [12, 13, 27], or simplified enacted high-quality caregiver speech (e.g., using CAREGIVER corpus [28], as in [14–17]). This leaves it unclear whether the audiovisual learning strategy also scales up to the real-world experiences of human infants, for whom speech and visual experience may be much more unconstrained than in the idealized speech corpora.

The goal of the present paper is to take a step towards an ecologically plausible test of the audiovisual bootstrapping hypothesis. We first introduce a novel neural network model for weakly-supervised word learning from acoustic speech and utterance-level labels that simulate visual attention of the language-learner. The model is then applied to data from real infant-caregiver interactions, as captured by head-mounted cameras and microphones worn by the infants, and tested on its capability to learn words from such experiences. In addition, we evaluate how latent representations (hidden unit activations) in the model relate to phonetic categories in input speech.

## 2. Methods

### 2.1. Theoretical background

The intuitive motivation for using non-linguistic multimodal input in bootstrapping of lexical learning comes from the fact that any acoustic word form has to have a meaning attached to it before it becomes a linguistic symbol and therefore plays any useful role in speech comprehension or production. In the same way, phonemic categories are defined in terms of semantic distinctions, which, by definition, requires at least some sort of lexical knowledge in the given language. On the other hand, first segmenting words and then trying to find their referents risks suboptimal segmentation strategies, as segmentation in isolation is not a well-defined task. Therefore, it would make most sense to optimize the entire representational system in a manner that provides the maximal information on the external world whenever speech is heard.

To formulate the same idea mathematically, it was proposed in [11] that the overall quality $Q$ of a (learned) lexicon $L$ can be defined as

$$Q = \sum_{w,c} P(w,c) \log_2 \frac{P(w,c)}{P(w)P(c)} / \max(\log_2|C|, \log_2|L|) \quad (1)$$

where $c \in C$ is the set of possible referents, $w \in L$ are the words of the lexicon, $P(x, y)$ stands for a joint probability distribution, and $|L|$ and $|C|$ denote the number of unique words and referents, respectively. In other words, the best lexicon is the one that maximizes the mutual information (MI) between words and referents (i.e., how many bits one knows about the external world, given speech input), constrained by the size of the lexicon to avoid redundant many-to-one mappings. Since young infants do not yet know the words $w$, and since the abstract categorical words $w$ are generally not directly observable from speech acoustics, Eq. (1) can be rewritten as

$$Q = \sum_{X,c} P(X,c|\theta) \log_2 \frac{P(X, c|\theta)}{P(X|\theta)P(c)} / \log_2|C| \quad (2)$$

where $X$ is acoustic speech input and $\theta$ defines an acoustic model that captures the joint density of speech $X$ with each possible referent $c$. In this case, a model $\theta^*$ that optimizes the MI between observable speech and referents in the communicative environment of the learner also leads to the best vocabulary in terms of referential value [11]. By learning such a model, word boundaries emerge as a side product from the process as points in time where the predictive distribution $P(c_t | X_t, \theta)$ changes from a set of referential predictions to another (see [11] for examples).

In this paper, we formulate the learning problem as cross-entropy minimization between latent representations derived from speech, and the visual objects attended by the infant, and explore how this strategy works out in practice. We also hypothesize that phonetic knowledge could emerge from the same optimization process as latent representations mediating the conversion from speech acoustics to word meanings, thereby also removing the need for separate phonetic learning "module" or "task" in the learner's brain.

### 2.2. Model implementation with deep neural networks

In order to implement the theoretical model using a flexible and powerful machine learning framework that would be also compatible with the hierarchical organization of sensory processing in the brain, deep neural networks were utilized. The basic modeling assumption here is that caregivers will occasionally speak utterances that contain words related to visual objects and events attended by the learner, and the learner utilizes those learning scenarios by updating a statistical model.

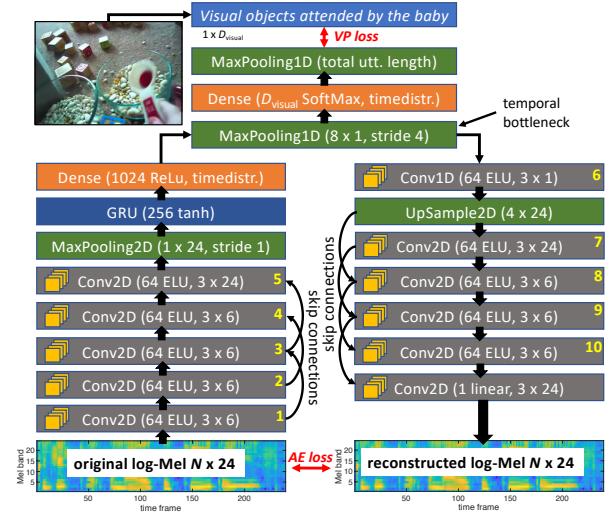

Figure 1: *CRNN architecture for cross-situational word learning. Numbers on convolutional layers correspond to the phone selectivity analysis layers in Fig 3.*

This means that the basic goal of the model is to learn acoustic patterns that predict the corresponding visual objects when the visual targets are only aligned at the utterance, but not at the word, level. Our solution to this problem is based on a convolutional recurrent neural network (CRNN) (Fig. 1). The CRNN network uses convolutional layers to encode a hierarchy of increasingly complex spectrotemporal patterns, a recurrent layer to capture temporal structure, and temporal max-pooling across the entire utterance to solve the temporal ambiguity issue (e.g., [29]). A visual prediction (VP) loss and an auxiliary autoencoder (AE) loss are used to guide the learning process.

In detail, input to the model consists of log-Mel spectrograms ($N$ frames x 24 dim, 25-ms window with 10-ms frame shifts) that are processed by a stack of five 2D convolutional layers. Each layer consists of $C$ channels with exponential linear unit (ELU) activations [30], with deeper layers capturing increasingly complex time-frequency patterns of the input. The convolutional stack is followed by max pooling across the Mel-channels, and a recurrent gated unit (GRU) layer models temporal evolution of the convolution stack output. GRU output is then downsampled by factor of 4 using a maxpooling with length of 8 frames (80 ms), producing a temporal bottleneck that enforces temporal contiguity to GRU outputs (cf. temporal *jitter* used in [31]). Next, GRU output is fed to a fully-connected time-distributed ReLu layer, which then feeds to parallel paths: towards visual predictions (VP), and towards decoder part of the AE for reconstruction of the original input.

In the VP pipeline, ReLu outputs are fed to a time-distributed softmax layer with $d_{visual}$ units that produces a probability estimate for each visual pattern for each time frame (40-ms per frame). These probabilities are then max-pooled across the entire duration of the utterance. During training, the resulting 1 x $d_{visual}$ vector is then used to calculate loss w.r.t. one-hot encoded objects attended by the infant, as labeled by human annotators (see Section 3.1). The remaining decoder section of the AE is an approximate inverse of the convolutional stack in the encoder side, but now paired with convolutional

pre- and post-filters. Its final output of size $N \times 24$ is used to calculate AE loss w.r.t. the original input.

The motivation for including autoencoding branch in the model comes from the limitation of temporal max-pooling required for the visual predictions, as max-pooling across *all* frames in the input means that only a subset of all network nodes is updated for each input utterance. By including the AE loss, the model has to discover latent high-level representations that are suitable for both input spectrum reconstruction and prediction of the visual objects, and where AE loss optimization affects all parameters of the AE network for all input. As demonstrated in the experiments, this results in much higher performance compared to a network without the AE branch.

## 3. Experiments

### 3.1. Data

We used audiovisual SEEDLingS corpus [32,33] as the primary source of data for our experiments. The original dataset consists of videos with audio tracks from head-mounted cameras worn by infants when they interact with their caregivers at their own homes (without the presence of any outsider). Each of the babies in SEEDLingS were recorded for 1 hour at a time and every 1 month from the age of 6 months up to 14 months. So far, the data corresponding to 6- and 7-month-olds has been hand-labeled for all clearly spoken utterances containing concrete imageable nouns and for information whether the mentioned object was present and attended by the infant (yes/no) [33].

For our experiments, we extracted all annotated utterances from the 6- and 7-month data (43 different babies). All regular plurals were first merged with their singular forms (e.g., "*blocks*"→"*block*"), and then we chose all utterances containing nouns with at least 50 occurrences in all data as our training and testing data, leading to a set of 60 visually perceivable nouns. After pooling the audio data from all babies together, the total amount of utterances was 6650 with an average duration of 1.94 (±1.2) seconds (total 3h 35min). Note that this includes a large variety of voices (at least 43 unique talkers), speaking styles (normal, whisper, singing etc.) and auditory environments (background tv, air conditioning, overlapping sound from the target child, siblings, pets etc.). A training set was defined by randomly choosing 80% of utterances for each labeled noun (N = 6765 utterances). Rest of the data, were left for testing (N = 1353). For every training utterance for which the baby did not attend to an object mentioned in the utterance (35% of all cases), the visual object label was randomly sampled from the list of 60 object classes. A randomly chosen subset of 20% of all training utterances was used as a validation set for early stopping of model training.

In addition, we report results for CAREGIVER Y2 UK corpus (from now on "CG"; [28]) in order to ensure that the CRNN model performs correctly on data with established results for audiovisual learning (e.g., [11, 15, 17]). In our experiments, we use data from the four primary talkers of the corpus, each speaking 2397 English utterances in enacted child-directed speaking style. The utterances consists of carrier sentences (function words etc.) and 1–4 so-called keywords that come with a related unordered bag of visual labels, simulating the concurrent presence of corresponding visually perceivable objects and actions. All CG data have been recorded in high-quality sound-isolated room. In our experiments, we randomly divided 80% of the data into training (N = 7670 utterances, 20% again used for validation), and left 20% (N = 1918) for testing.

### 3.2. Experimental setup

Adam optimizer ($\alpha = 0.01$, $\beta_1 = 0.9$, $\beta_2 = 0.999$), early stopping with patience of 10, and minibatch size of 50 utterances were used for model training. The number of convolution kernels per layer was set to $C = 64$, GRU units to 256, and time distributed ReLu units to 1048. Categorical cross-entropy was used for VP loss (weighted with the inverse frequency of each noun/object to compensate for a highly skewed distribution) and RMSE for AE loss. All log-Mel inputs were zero-padded or clipped to 512 frames (95% of all utterances were shorter than 512 frames). Dropout rate of 0.1 was used after all convolutions in the encoder, after GRU output, and in the ReLu layer, and L2-regularization ($\lambda = 0.01$) was applied to the weights of all but GRU and softmax layers. All experiments were conducted with Keras using TensorFlow backend.

To demonstrate the importance of AE part of our model, we also report results for otherwise identical model but only trained in the visual prediction loss ("*no-AE*"), and for a model without the temporal bottleneck ("*AE-noBN*"). We also tested a variant of the model where the decoder, instead of reconstructing the original input, tries to predict input spectrum 250-ms (approx. one syllable) ahead ("*AE-pred*"), and a variant where the AE branch is first pre-trained to early stopping before jointly optimizing the entire network for AE and VP ("*AE-pretrain*"). Since the pretraining variant did not lead to good performance, it is not reported in the results. As a sanity check baseline, we also report results for the so-called CM-algorithm [15], a high-order Markov chain approximation of Eq. (2), which represents typical word learning performance on the CG corpus among the models applied to the same task earlier (see also, e.g., [17]).

The success of word learning was evaluated as a word detection task: a spoken word was considered as detected if the probability of the corresponding visual object, given an utterance containing the name of the object, exceeded a detection threshold $\gamma$. Recall (proportion of true targets detected), precision (proportion of detections correct), and F-score (harmonic mean of the previous two) are reported as a function of $\gamma$. Due to the skewed word frequency distribution, the overall performance is reported as the mean of word-specific measures ("unweighted" metrics). All model variants were trained three times with different random seeds, and the average across all runs is reported for each model variant.

In addition, we analyzed the selectivity of hidden layer activations towards different phone categories using English utterances of from the test section of TIMIT corpus [34] using reduced TIMIT phone set (39 phone categories) and the best performing model trained on the SEEDLingS data. More specifically, phone selectivity index (PSI) [35, 36] was computed for each layer. In short, a node- and phone-specific PSI value describes the average number of phones (0–38) whose activation patterns are distinct from the node's activation pattern during the given phone (tested with unpaired t-test, p < 0.05), higher PSI meaning better separability of phone categories. We report PSIs as averages across all nodes of a given layer, averaged across all phone categories in the input. T-test *t*-statistics from PSI calculation are also reported in order to quantify the average effect size of phone separability, also as a function of layer in the network. Selectivity analyses were conducted for model activations obtained during the middle frame of each annotated phone.

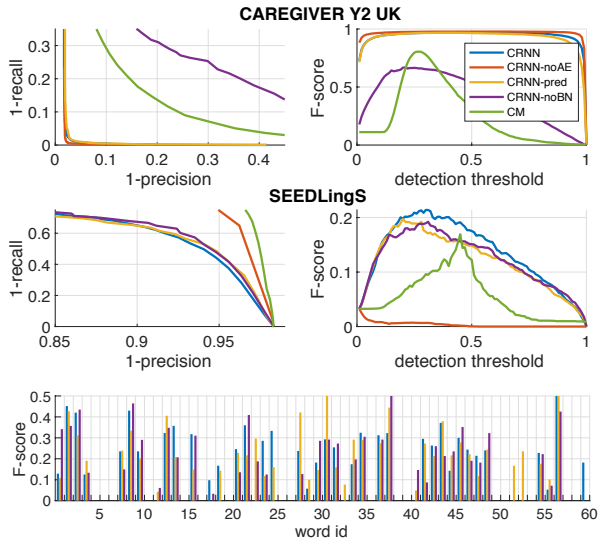

Figure 2: *Word detection performance on CAREGIVER (top) and SEEDLingS (middle) test sets as a function of word detection threshold γ. Different CRNN model variants and the CM baseline are shown with different colors. Word-specific F-scores on SEEDLingS are shown at the bottom for CRNN, CRNN-pred, and CRNN-noBN (γ = 0.3).*

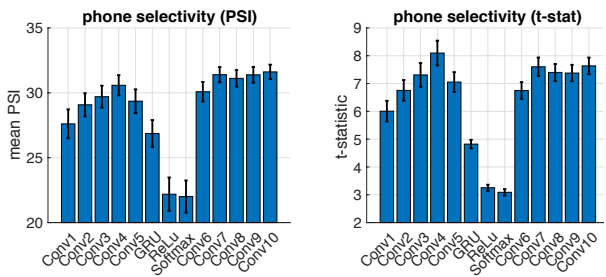

Figure 3: *Selectivity of CRNN network nodes in different hidden layers towards phone categories of TIMIT after weakly-supervised training on SEEDLingS audiovisual data. Error bars denote standard error across different phone categories. Higher bars correspond to higher phonetic selectivity.*

### 3.3. Results

We first verified on the studio-recorded CG corpus that the models are working correctly (Fig. 2, top). CRNN, CRNN-noAE, and CRNN-pred reach F-scores of ≥ 0.96 while CM has a baseline F-score of 0.81 with an optimal γ (Fig. 2, top panels). The CRNN variant without the temporal bottleneck (CRNN-noBN) performs worse than the baseline.

As for the results on the much more challenging SEEDLingS data from real infant-caregiver interactions (middle panel, Fig. 2), the pattern of results changes somewhat: The use of AE loss is now essential for successful learning, and the CRNN-noAE basically fails to converge to a solution on all three runs of the training. The basic CRNN outperforms the variants without temporal bottleneck (CRNN-noBN) and with temporal prediction AE (CRNN-pred) with a small margin.

Since the overall performance is not very high, bottom panel in Fig. 2 shows the word detection F-score for each of the 60 nouns in the SEEDLingS test set. As can be observed, some of the words are not learned at all (e.g., *milk, ring, or water*), while some others are detected with an F-score up to 0.4–0.5 (e.g., *book, car, ball, baby, cat,* and *doggy*). Furthermore, the confusions made by the model are not fully random. For instance, *bottle* is often confused with *ball* (54%), *cat* with *hat* (12%), *cow* with *car* (15%) *doggy* with *dog* (19%), and *foot* with *book* (11%). There are also some semantic-like confusions such as *nose* with *mouth* (27%), *tummy* with *knee* (24%), and *knee* with *toe* (76%) that may originate from the typical verbal contexts of object naming (numbers are CRNN means across three re-runs of the model; chance level for confusions is 1.7%).

Fig. 3 shows the selectivity of CRNN network layers towards phone categories of TIMIT after training on SEEDLingS. A pattern of increasing phonetic (~phonemic) selectivity in deeper convolutional layers can be observed, being similar to the gradually increasing selectivity found in feedforward networks of trained for phone recognition in a supervised manner [35]. Interestingly, selectivity is highest level throughout the decoder part of the autoencoder.

## 4. Discussion and conclusions

We interpret the result as a generally positive finding: the proposed model shows indications of successful learning of early proto-lexical and phonemic representations from audiovisual scenarios of real language-learning infants. Moreover, the learning takes place with only some hours of input, and succeeds in the task without prior linguistic knowledge or specialized mechanisms for phonemic learning or word segmentation.

However, it is still important to remember that the present learning setup is still simplified from reality. Only utterances with concrete nouns were included in the input, as those are only currently annotated for the SEEDLingS data. However, we did include all cases with erroneous attention in the training as "false" targets. In addition, the visual input was assumed to be categorical and invariant in nature. On the other hand, the present test setup is a fully unconstrained detection task from speech by tens of acoustically distinct talkers, very few word tokens per talker, and generally poor signal-to-noise ratio mono recordings. In contrast, real infants would typically hear speech from a small number of primary caregivers, and would be judged in their performance in more constrained communicative contexts (e.g., orienting towards a [cup] instead of [mommy] when hearing "*cup*"). In addition, the age-range of the children in the dataset is only 6–7 months, whereas typical US English children show (parent-reported) comprehension of approximately ten different words at the age of 8 months [37].

Technically, the proposed model demonstrates state-of-the-art performance on the GG corpus (cf. [17]), and also shows the benefit of autoencoder-based auxiliary loss in a weakly-supervised learning scenario. The present model should be also applicable to modeling of adult speech perception [38, 39] and phonemic categorization (cf. [35]), where the previous work has only utilized ecologically problematic supervised learning from speech to perfectly aligned semantic or phonetic targets.

As a next step in our research, our aim is to substitute the visual labels with the actual visual input captured by the head-mounted cameras on the infants. Only by doing so we can truly understand the richness and complexity of the audiovisual experiences available to the learners.

## 5. Acknowledgements

This research was funded by Academy of Finland grants no. 314602 and 320053. The authors would like to thank Elika Bergelson for her help with the SEEDLingS dataset.